\documentclass[11pt,a4paper]{article}
\usepackage[utf8]{inputenc}
\usepackage[hyperref]{emnlp2020}
\usepackage{times}
\usepackage{algorithm}
\usepackage{algpseudocode}
\usepackage{todonotes}
\usepackage{amsmath}
\usepackage{multirow}
\usepackage{graphicx}
\usepackage{subcaption}
\usepackage{enumitem}
\usepackage{booktabs}
\usepackage{CJKutf8}

\newcommand{\subtab}[2]{\begin{tabular}{@{}#1@{}} #2 \end{tabular}}
\newcommand{\ja}[1]{\begin{CJK*}{UTF8}{min} #1 \end{CJK*}}

\usepackage{microtype}

\aclfinalcopy % Uncomment this line for the final submission
\def\aclpaperid{3409} %  Enter the acl Paper ID here

\title{Byte Pair Encoding is Suboptimal for Language Model Pretraining}

\author{Kaj Bostrom \and Greg Durrett \\
  Department of Computer Science\\
  The University of Texas at Austin \\
  \texttt{\{kaj,gdurrett\}@cs.utexas.edu}}

\date{}

\begin{document}
\maketitle
\begin{abstract}

\hyphenpenalty=5000
\exhyphenpenalty=5000
The success of pretrained transformer language models (LMs) in natural language processing has led to a wide range of pretraining setups.
In particular, these models employ a variety of subword tokenization methods, most notably byte-pair encoding (BPE) \citep{sennrich-etal-2016-neural, Gage1994NAD}, the WordPiece method \citep{Schuster2012JapaneseAK}, and unigram language modeling \citep{kudo-2018-subword}, to segment text.
However, to the best of our knowledge, the literature does not contain a direct evaluation of the impact of tokenization on language model pretraining. We analyze differences between BPE and unigram LM tokenization, finding that the latter method recovers subword units that align more closely with morphology and avoids problems stemming from BPE's greedy construction procedure. We then compare the fine-tuned task performance of identical transformer masked language models pretrained with these tokenizations. Across downstream tasks and two languages (English and Japanese), we find that the unigram LM tokenization method matches or outperforms BPE. We hope that developers of future pretrained LMs will consider adopting the unigram LM method over the more prevalent BPE.
\end{abstract}

\section{Introduction}
Large transformers \citep{vaswani2017attention} pretrained with variants of a language modeling objective, such as \textsc{BERT} \citep{devlin2019bert}, have proven their effectiveness at flexibly transferring to a variety of domains and tasks. One design decision that makes them particularly adaptable is their graceful handling of the open vocabulary problem through subword tokenization. Subword tokenization, popularized in the neural machine translation literature \citep{sennrich-etal-2016-neural, vaswani2017attention, wu2016google}, produces tokens at multiple levels of granularity, from individual characters to full words. As a result, rare words are broken down into a collection of subword units, bottoming out in characters in the worst case.

Critically, a pretrained language model's subword vocabulary cannot be altered: any downstream application of these models must tokenize input or generate output using the original subword vocabulary, making the choice of tokenization a particularly significant decision.

A variety of subword tokenization methods have seen use in pretrained language models. \textsc{BERT} uses the WordPiece method \citep{Schuster2012JapaneseAK}, a language-modeling based variant of BPE;
T5 \citep{raffel2019exploring} uses character-level BPE;
\textsc{GPT2} \citep{radford2019language} and \textsc{RoBERTa} \citep{liu2019roberta} use BPE over raw bytes instead of unicode characters;
\textsc{XLNet} \citep{yang2019xlnet} and \textsc{ALBERT} \citep{lan2019albert} use the SentencePiece library \citep{kudo-richardson-2018-sentencepiece} which implements both BPE and unigram language model tokenization, but in both cases fail to clarify which of these methods they chose. The effects of tokenization are not examined in a reported experiment in any of the above works except \citet{liu2019roberta}, who note that WordPiece gave a small advantage over BPE in their preliminary investigation. In the machine translation literature, \citet{kudo-2018-subword} introduced the unigram language model tokenization method in the context of machine translation and found it comparable in performance to BPE. \citet{domingo2018much} performed further experiments to investigate the effects of tokenization on neural machine translation, but used a shared BPE vocabulary across all experiments. \citet{galle-2019-investigating} examined algorithms in the BPE family, but did not compare to unigram language modeling.

%The effects of tokenization are not examined in a reported experiment in any of the above works, excepting \citet{liu2019roberta} who note that ``Early experiments revealed only slight differences between \lbrack WordPiece and BPE\rbrack, with the \citet{radford2019language} BPE achieving slightly worse end-task performance on some tasks \lbrack than the \citet{devlin2019bert} WordPiece\rbrack .''

%\citet{kudo-2018-subword} investigate the impact of subword tokenization on neural machine translation, introducing the unigram language model tokenization method. They conduct a comparison to BPE tokenization for English-German translation, finding their performance comparable. \citet{domingo2018much} perform further experiments to investigate the effects of tokenization on neural machine translation, however they use a shared BPE vocabulary across all experiments.
 
In this work, we characterize the space of proposed subword tokenization algorithms and analyze the differences between the two methods with publicly available implementations: BPE (merging tokens based on bigram frequency) and unigram language modeling (pruning tokens based on unigram LM perplexity). While the vocabularies resulting from these schemes are heavily overlapping, we compare each method to reference morphological segmentations and find that the unigram LM method produces tokens better aligned with morphology. To understand whether this more natural tokenization leads to improved performance, we pretrain separate language models using the \textsc{RoBERTa} objective \cite{liu2019roberta} with each tokenization for both English and Japanese, two typologically distant languages. On downstream tasks, we find a performance gap across tasks and languages, with the unigram LM method providing an improvement over BPE of up to 10\% in our Japanese QA experiments, indicating the benefits of adopting this technique in the context of language model pretraining.

\section{Algorithms}

Subword tokenization algorithms consist of two components: a vocabulary construction procedure, which takes a corpus of text and returns a vocabulary with the desired size, and a tokenization procedure, which takes the built vocabulary and applies it to new text, returning a sequence of tokens. In theory, these two steps can be independent, although for the algorithms we examine the tokenization procedure is tightly coupled to the vocabulary construction procedure.

A BPE vocabulary is constructed as follows:

\begin{algorithm}
\caption{Byte-pair encoding \citep{sennrich-etal-2016-neural, Gage1994NAD}}\label{bpe}
\begin{algorithmic}[1]
\State Input: set of strings $D$, target vocab size $k$
\Procedure{BPE}{$D, k$}
\State $V \gets$ all unique characters in $D$
\State \hspace{2em} (about 4,000 in English Wikipedia)
\While{$|V| < k$} \Comment{Merge tokens}
    \State $t_L, t_R \gets $ Most frequent bigram in $D$
    \State $t_\textsc{new} \gets t_L + t_R$ \Comment{Make new token}
    \State $V \gets V + \lbrack t_\textsc{new} \rbrack$
    \State Replace each occurrence of $t_L, t_R$ in
    \State \hspace{1em} $D$ with $t_\textsc{new}$
\EndWhile
\State \textbf{return} $V$
\EndProcedure
\end{algorithmic}
\end{algorithm}

BPE tokenization takes the vocabulary $V$ containing ordered merges and applies them to new text in the same order as they occurred during vocabulary construction.

The WordPiece algorithm \citep{Schuster2012JapaneseAK}, used to construct BERT's vocabulary, closely resembles BPE. However, instead of merging the most frequent token bigram, each potential merge is scored based on the likelihood of an $n$-gram language model trained on a version of the corpus incorporating that merge.
\citet{Schuster2012JapaneseAK} note that the process of estimating language model parameters for every potential merge is prohibitive, so they employ aggressive heuristics to reduce the number of potential merges considered. As their implementation is not public,\footnote{Although its name and association with Google might suggest otherwise, the SentencePiece library \citep{kudo-richardson-2018-sentencepiece} does not, in fact, implement the WordPiece algorithm; it provides implementations of BPE and unigram LM based tokenization.} we are unable to make a comparison to this method.

The unigram LM method \citep{kudo-2018-subword}, in contrast to the bottom-up construction process of BPE and WordPiece, begins with a superset of the final vocabulary, pruning it to the desired size:
\begin{algorithm}
\caption{Unigram LM \citep{kudo-2018-subword}}\label{unigram}
\begin{algorithmic}[1]
\State Input: set of strings $D$, target vocab size $k$
\Procedure{UnigramLM}{$D, k$}
\State $V \gets$ all substrings occurring more than
\State \hspace{2em} once in $D$ (not crossing words)
\While{$|V| > k$} \Comment{Prune tokens}
    \State Fit unigram LM $\theta$ to $D$
    \For{$t \in V$} \Comment {Estimate token `loss'}
        \State $L_t \gets p_{\theta}(D) - p_{\theta'}(D)$
        \State where $\theta'$ is the LM without token $t$
    \EndFor
    \State Remove $\min (|V|-k, \lfloor \alpha |V| \rfloor )$ of the 
    \State tokens $t$ with highest $L_t$ from $V$,
    \State where $\alpha \in \lbrack0, 1\rbrack$ is a hyperparameter
\EndWhile
\State Fit final unigram LM $\theta$ to $D$
\State \textbf{return} $V, \theta$
\EndProcedure
\end{algorithmic}
\end{algorithm}

\begin{figure*}[!t]
\small
    \centering
    \begin{tabular}{r l}
        \textbf{Original:} & furiously \\
        \textbf{BPE:} & \texttt{\_fur iously} \\
        \vspace{1ex}\textbf{Uni. LM:} & \texttt{\_fur ious ly} \\
    \end{tabular}
    \begin{tabular}{r l}
        \textbf{Original:} & tricycles \\
        \textbf{BPE:} & \texttt{\_t ric y cles} \\
        \vspace{1ex}\textbf{Uni. LM:} & \texttt{\_tri cycle s} \\
    \end{tabular}
    \begin{tabular}{r l}
        \textbf{Original:} & nanotechnology \\
        \textbf{BPE:} & \texttt{\_n an ote chn ology} \\
        \vspace{1ex}\textbf{Uni. LM:} & \texttt{\_nano technology}
    \end{tabular}
    \begin{tabular}{r l}
        \textbf{Original:} & \texttt{Completely preposterous suggestions} \\
        \textbf{BPE:} & \texttt{\_Comple t ely \_prep ost erous \_suggest ions} \\
        \vspace{1ex}\textbf{Unigram LM:} & \texttt{\_Complete ly \_pre post er ous \_suggestion s}
    \end{tabular} \\
    \begin{tabular}{l r}
        \begin{tabular}{@{}r l}
            \textbf{Original:} & \texttt{corrupted} \\
            \textbf{BPE:} & \texttt{\_cor rupted} \\
            \vspace{1ex}\textbf{Unigram LM:} & \texttt{\_corrupt ed}
        \end{tabular}
        & 
        \begin{tabular}{r l@{}}
            \textbf{Original:} & \texttt{1848 and 1852,} \\
            \textbf{BPE:} & \texttt{\_184 8 \_and \_185 2,} \\
            \vspace{1ex}\textbf{Unigram LM:} & \texttt{\_1848 \_and \_1852 ,}
        \end{tabular}
    \end{tabular} \\
    %\begin{tabular}{r l l l}
    %    \textbf{Original} & \multicolumn{3}{l}{\ja{戦争の激化}} \\
    %    \textbf{BPE} & \multicolumn{3}{l}{\ja{戦争の　激　化}} \\
    %    \textbf{Unigram LM} & \ja{戦争} & \ja{の} & \ja{激化} \\
    %    \textbf{Gloss} & war & of & intensification \\
    %    \vspace{1ex}\textbf{Translation} & \multicolumn{3}{l}{Intensification of war}
    %\end{tabular} \\
    \begin{tabular}{r l l l l l l l}
        \textbf{Original} & \multicolumn{7}{l}{\ja{磁性は様々に分類がなされている。}} \\
        \textbf{BPE} & \multicolumn{2}{l}{\ja{磁　性は}} & \ja{様々} & \multicolumn{2}{l}{\ja{に分類}} & \ja{がなされている} & \ja{。} \\
        \textbf{Unigram LM} & \ja{磁　性} & \ja{は} & \ja{様々} & \ja{に} & \ja{分類} & \ja{がなされている} & \ja{。}\\
        \textbf{Gloss} & magnetism & (top.) & various ways & in & classification & is done & . \\
        \textbf{Translation} & \multicolumn{6}{l}{Magnetism is classified in various ways.}
    \end{tabular}
    
    \caption{Example tokenizations. The character `\_' is a word boundary marker. BPE merges common tokens, such as English inflectional suffixes and Japanese particles, into their neighbors even when the resulting unit is not semantically meaningful.}
    \label{fig:examples}
\end{figure*}

\begin{figure*}[!t]
\small
    \centering
    \begin{subfigure}[b]{\columnwidth}
        \includegraphics{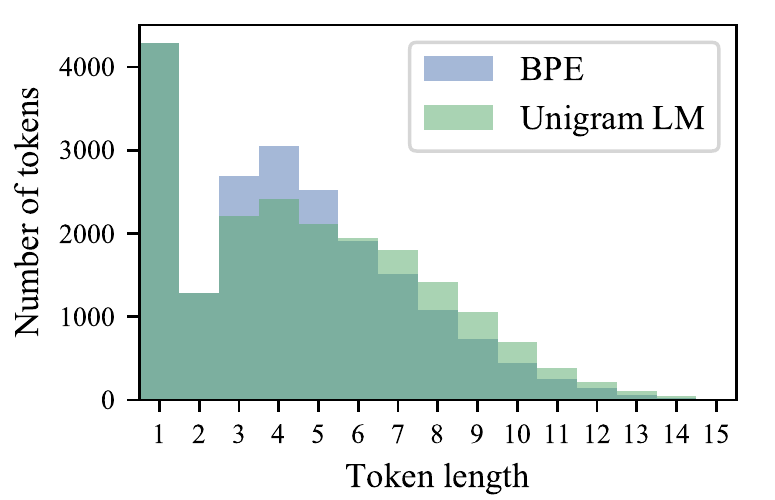}
        \caption{Token length distributions within each vocabulary}
        \label{fig:token_lengths}
    \end{subfigure}
    \begin{subfigure}[b]{\columnwidth}
        \includegraphics{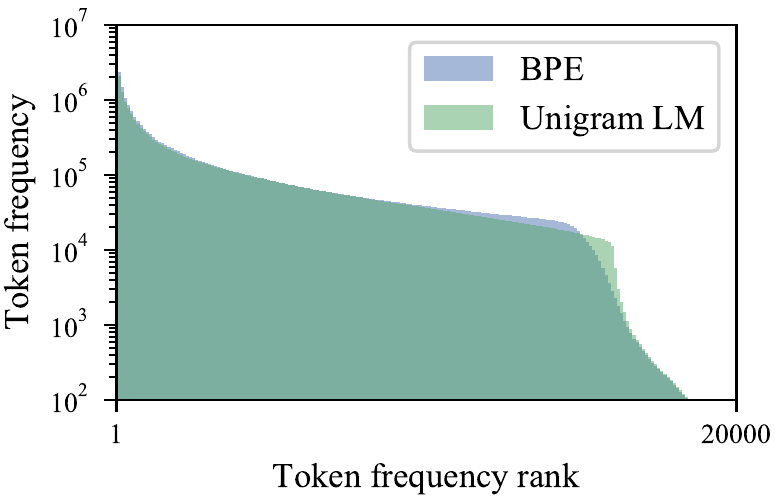}
        \caption{Token frequency profiles over the corpus}
        \label{fig:token_freqs}
    \end{subfigure}
    \caption{English subword vocabulary and corpus profiles. The unigram LM method produces longer tokens on average (a) and uses its vocabulary space more effectively (b), with more tokens of moderate frequency.}
\end{figure*}

Unigram LM tokenization takes the vocabulary $V$ and unigram LM parameters $\theta$ and performs Viterbi inference to decode the segmentation with maximum likelihood under $\theta$. This method is similar to Morfessor's unsupervised segmentation \cite{creutz2005unsupervised} without its informed prior over token length.

In the course of our experiments we did not observe a major difference in speed between the two algorithms. Both require similar amounts of time to construct a vocabulary, and both have a negligible impact on overall model inference latency.

\section{Comparison of Segmentations}

\subsection{Morphology}

In Figure \ref{fig:examples} we illustrate the differences in tokenization output between BPE and the unigram LM method. We observe that the unigram LM method produces subword units that qualitatively align with morphology much better than those produced by BPE.
In particular, we note that the unigram LM method recovers common affixes such as \emph{-ly}, \emph{-s}, \emph{pre-}, and \emph{tri-} while BPE does not, instead absorbing them into adjacent units (\emph{-cles}) while also producing meaningless single-character units.

\begin{table}[!t]
\small
    \centering
    \begin{tabular}{c c} \toprule
        \multicolumn{2}{c}{\textbf{More frequent in}} \\
        BPE & Unigram LM \\ \midrule
        \begin{tabular}{@{}c c c c c}
        \_H & \_L & \_M & \_T & \_B \\ 
        \_P & \_C & \_K & \_D & \_R
        \end{tabular} &
        \begin{tabular}{c c c c c@{}}
        s & . & , & ed & d \\
        ing & e & ly & t & \_a
        \end{tabular} \\
        \bottomrule
    \end{tabular}
    \caption{Tokens with the highest difference in frequency between tokenizations. The unigram LM method tends to produce more parsimonious prefixes and suffixes.}
    \label{tab:max_diff}
\end{table}

\begin{table}[!t]
\small
    \centering
    \begin{tabular}{l c c} \toprule
         & \multicolumn{2}{c}{\textbf{Tokenization}} \\
         & BPE & Unigram LM \\ \midrule
        Tokens per word type & 4.721 & 4.633 \\
        Tokens per word & 1.343 & 1.318 \\ \bottomrule
    \end{tabular}
    \caption{Mean subword units per word for each method across all of English Wikipedia.}
    \label{tab:vocab_stats}
\end{table}

\begin{table*}[!t]
\small
    \centering
    \begin{tabular}{l c c c c c c}\toprule
    \multirow{2}{*}{\textbf{Method}} & \multicolumn{3}{c}{\textbf{English (w.r.t. CELEX2)}} & \multicolumn{3}{c}{\textbf{Japanese (w.r.t. MeCab)}} \\
    & Precision & Recall & F1 & Precision & Recall & F1 \\ \midrule
    BPE & 38.6\% & 12.9\% & 19.3\% & 78.6\% & 69.5\% & 73.8\% \\
    Uni. LM & 62.2\% & 20.1\% & 30.3\% & 82.2\% & 72.8\% & 77.2\% \\ \bottomrule
    
    \end{tabular}
    \caption{Correspondence of subword boundaries between unsupervised tokenization methods and morphological reference segmentations.}
    \label{tab:quant_morph}
\end{table*}

This trend is supported by Table \ref{tab:max_diff}, in which we observe that recognizable affixes appear much more frequently in the unigram LM tokenization of our pretraining corpus than in the BPE tokenization.
As the BPE tokenization is constructed greedily according to frequency, common affixes (and punctuation) are frequently absorbed into other tokens.\footnote{Note that the BPE vocabulary still includes these affixes, but when they are encountered during tokenization, they are almost always merged into larger units as in Figure~\ref{fig:examples}.}

We see in Figure \ref{fig:token_lengths} that the unigram LM tokenization tends to have longer subword units than BPE. This is closer to the length distribution of gold-standard English morphs, which have a mean length of approximately 6 characters \citep{creutz2004morpheme}. 

\paragraph{Comparison with morphological segmenters} In Table \ref{tab:quant_morph}, we further corroborate these observations by performing a quantitative evaluation of the degree to which each unsupervised segmentation algorithm aligns with morphological baselines for each language. For English, we produce gold surface allomorph boundaries from the CELEX2 lexical database \citep{celex2} in the manner of \citet{hutmegs}. We then compare each algorithm's subword unit boundaries with gold morpheme boundaries for words with 2 or more morphemes, weighted by their frequency in English Wikipedia. For Japanese, we compare subword tokenizations of Japanese Wikipedia sentences to morphological reference tokenizations produced using the MeCab morphological analysis and tokenization tool \citep{mecab} using version 2.3.0 of the UniDic dictionary \citep{unidic}.

We find that for both languages, the segmentations produced by the unigram LM method correspond more closely to the morphological references, confirming our qualitative analysis. On English data, both unsupervised methods exhibit low boundary recall; we attribute this to the fact that they represent many common words with underlying derivational morphology as single tokens, although for BPE this is compounded by effects we discuss in Section 3.2.

The ability of the unigram LM method to recover the morphological structure of the text without explicit supervision aligns with the main findings of \citet{creutz2005unsupervised}, who successfully use maximum-a-posteriori unigram language models to perform unsupervised morphological segmentation of English and Finnish.

\subsection{Vocabulary Allocation}

By surfacing subword units that align with morphology, the unigram LM tokenization provides the opportunity for the model to learn composable subword embeddings.
If an affix reliably signals a linguistic feature, rather than needing to store that information redundantly across the embeddings of many tokens containing the affix, the model can store it in just the embedding of the affix.

\begin{table*}[!t]
\small
    \centering
    \begin{tabular}{l |c c c c c c| c c} \toprule
        & \multicolumn{6}{|c|}{\textbf{English}} & \multicolumn{2}{c}{\textbf{Japanese}}\\
        \multirow{2}{*}{\textbf{Model}} & \multicolumn{2}{|c}{\textbf{SQuAD 1.1} (dev.)} & \multicolumn{2}{c}{\textbf{MNLI} (dev.)} & \multicolumn{2}{c|}{\textbf{CoNLL NER}} &
        \multicolumn{2}{c}{\textbf{TyDi QA} (dev.)} \\
        & EM & F1 & Acc. (m) & Acc. (mm) & Dev. F1 & Test F1 & EM & F1 \\
        \midrule
        Ours, BPE & $80.6\pm.2$ & $88.2\pm.1$ & $81.4\pm.3$ & $82.4\pm.3$ & $94.0\pm.1$ & $90.2\pm.0$ & $41.4\pm0.6$ & $42.1\pm0.6$ \\
        Ours, Uni. LM & $81.8\pm.2$ & $89.3\pm.1$ & $82.8\pm.2$ & $82.9\pm.2$ & $94.3\pm.1$ & $90.4\pm.1$ & $53.7\pm1.3$ & $54.4\pm1.2$ \\
        \midrule
        BERT$_\textsc{base}$ & 80.5 & 88.5 & 84.6 & 83.4 & 96.4 & 92.4 & -- & -- \\
        \bottomrule
    \end{tabular}
    \caption{Fine-tuning results. Metrics are averaged across 5 fine-tuning seeds with standard deviations indicated by $\pm$; due to computational constraints we did not pretrain more than once per tokenization. We include fine-tuning results for a transformer with a comparable architecture, BERT$_\textsc{base}$, for reference, although we note that a direct comparison cannot be made due to BERT$_\textsc{base}$ using both a larger pretraining corpus and a larger subword vocabulary.}
    \label{tab:fine-tune}
\end{table*}

These results suggest that the unigram LM method may allocate its vocabulary more economically. We note in Figure \ref{fig:token_freqs} that both vocabularies contain a ``dead zone'' of tokens whose frequency is much lower than the rest of the vocabulary. This is largely the result of the presence of a number of very uncommon characters, including Chinese and Japanese kanji, in the training corpus. In the BPE tokenization, however, this effect is exacerbated, with the dead zone containing about 1500 more entries as a result of the tendency of its vocabulary construction process to produce intermediate ``junk'' tokens. For example, in the case where three tokens almost always occur as a group, in order to merge them into a single token, BPE must first merge one pair before incorporating the third token; this leaves an intermediate token in the vocabulary that will only occur rarely on its own. Additionally, tokens that appear in many contexts, such as inflectional affixes (-s, -ed), will tend to merge with many adjacent units due to their frequency. However, these merges lead to embedding redundancy, as these affixes usually have the same linguistic function in every context. Since the unigram LM method selects tokens during vocabulary construction using a global optimization procedure, it does not produce junk tokens; this property also allows it to avoid merging frequent tokens with their neighbors too aggressively.

Japanese vocabulary comparisons are included in Appendix B.

\section{Downstream Task Experiments}

In order to make a fair experimental comparison between these two methods on downstream tasks, we do not use an existing pretrained language model like BERT, but instead train our own language models from scratch, controlling for the data, training objective, and optimization procedure.
We pretrain four transformer masked language models using the architecture and training objective of \textsc{RoBERTa-base} \citep{liu2019roberta} using the reference \texttt{fairseq} implementation \citep{ott-etal-2019-fairseq}. Two are pretrained on the text of English Wikipedia, comprising $\sim$3B tokens under either tokenization. The other two are pretrained on the text of Japanese Wikipedia, comprising $\sim$0.6B tokens.
In each pair, one model is pretrained on the BPE tokenization of the corpus, and the other on the unigram LM tokenization, each with a vocabulary of 20,000 tokens. Hyperparameters are listed in Appendix A.

We subsequently fine-tune each of the pretrained English models on the SQuAD question-answering task \citep{rajpurkar-etal-2016-squad}, the MNLI textual entailment task \citep{williams-etal-2018-broad}, and the English portion of the CoNLL 2003 named-entity recognition shared task \citep{tjong-kim-sang-de-meulder-2003-introduction}. We fine-tune the Japanese models on the Japanese minimal-answer subset of the TyDi question-answering task \citep{clark2020tydi}. We base our fine-tuning implementations on those of the \texttt{transformers} toolkit \citep{wolf2019transformers}.

The results of our fine-tuning experiments are presented in Table \ref{tab:fine-tune}. We show that fine-tuning models pretrained with unigram LM tokenization produces better performance than fine-tuning models pretrained with BPE tokenization for all tasks. These results suggest that the higher morphological plausibility of the unigram LM tokenization may translate into better downstream task performance as well. Larger performance gaps are evident on SQuAD and MNLI, but the largest gap appears on Japanese TyDi. Differences in pretraining may be more evident in this setting due to the fact that the Japanese portion of the TyDi training split only contains $\sim$5k examples, compared to the $\sim$88k examples available for fine-tuning on SQuAD. Additionally, written Japanese does not feature whitespace between words, so it is possible for tokenizations to differ in word boundary placement as well as subword segmentation.

\section{Conclusion}

In this work we show that the choice of input encoding makes a difference in how well pretrained language models are able to perform end tasks. This indicates that tokenization encodes a surprising amount of inductive bias, and we suggest that unigram LM tokenization may be the better choice for development of future pretrained models.

\section*{Acknowledgments}

This work was partially supported by NSF Grant IIS-1814522 and a gift from Arm. This material is also based on research that is supported by the Air Force Research Laboratory (AFRL), DARPA, for the KAIROS program under agreement number FA8750-19-2-1003. The U.S. Government is authorized to reproduce and distribute reprints for Governmental purposes notwithstanding any copyright notation thereon. The views and conclusions contained herein are those of the authors and should not be interpreted as necessarily representing the official policies or endorsements, either expressed or implied, of the Air Force Research Laboratory (AFRL), DARPA, or the U.S. Government.

\interlinepenalty=10000

\bibliography{anthology,acl2020}
\bibliographystyle{acl_natbib}

\newpage

\interlinepenalty=4000

\appendix
\section{Hyperparameters}
\small{
\begin{tabular}{l l} \toprule
    \textbf{Pretraining} & \\
    \midrule
    Model architecture & \subtab{l}{\textsc{RoBERTa-base} \\ \citep{liu2019roberta}\vspace{1ex}} \\
    Implementation & \subtab{l}{\texttt{fairseq} \\ \citep{ott-etal-2019-fairseq}\vspace{1ex}} \\
    Optimizer & \subtab{l}{\textsc{Adam}, $\epsilon = 1\text{e-}6$ \\ $\beta = (0.9, 0.98)$ \\ \citep{kingma2014adam}\vspace{1ex}} \\
    Learning rate decay\vspace{0.25ex} & Polynomial \\
    Peak learning rate\vspace{0.25ex} & $0.0005$ \\
    Warmup steps\vspace{0.25ex} & $10000$ \\
    Weight decay\vspace{0.25ex} & $0.01$ \\
    Batch size\vspace{0.25ex} & $2048$ \\
    Sequence length\vspace{0.25ex} & $512$ \\
    Total updates\vspace{0.25ex} & $125000$ \\
    MLP dropout\vspace{0.25ex} & $0.1$ \\
    Attention dropout\vspace{0.25ex} & $0.1$ \\
    Precision\vspace{0.25ex} & 16-bit \\ \midrule
    \textbf{Fine-tuning} \\ \midrule
    Implementations & \subtab{l}{\texttt{transformers} \\ \citep{wolf2019transformers}\vspace{1ex}} \\
    Optimizer & \subtab{l}{\textsc{Adam}, $\epsilon = 1\text{e-}8$ \\ $\beta = (0.9, 0.999)$ \vspace{1ex}} \\
    Learning rate decay\vspace{0.25ex} & Linear \\
    Peak learning rate\vspace{0.25ex} & $5\text{e-}5$ \\
    Warmup steps\vspace{0.25ex} & $0$ \\
    Weight decay\vspace{0.25ex} & $0$ \\
    Batch size\vspace{0.25ex} & $32$ \\
    \subtab{l}{Sequence length\\ (SQuAD, TyDi QA)\vspace{0.25ex}} & $512$ \\
    \subtab{l}{Passage stride\\ (SQuAD, TyDi QA)\vspace{0.25ex}} & $192$ \\
    \subtab{l}{Sequence length\\ (MNLI, NER)\vspace{0.25ex}} & $128$ \\
    Epochs\vspace{0.25ex} & $3$ \\
    Precision\vspace{0.25ex} & 16-bit \\
    \midrule
    \textbf{Tokenization} \\
    \midrule
    Implementations\vspace{0.25ex} & \subtab{l}{SentencePiece \\ \citep{kudo-richardson-2018-sentencepiece}} \\
    Vocabulary size\vspace{0.25ex} & $20000$ \\
    Unigram LM $\alpha$ & $0.25$ \\
    \bottomrule
\end{tabular}
}

\begin{table*}[!h]
\section{Japanese vocabulary comparison}
\end{table*}

\begin{table*}[!h]
\small
    \centering
    \begin{tabular}{c@{\hspace{2em}} c} \toprule
        \multicolumn{2}{c}{\textbf{More frequent in}} \\
        BPE & Unigram LM \\ \midrule
        \begin{tabular}{@{}c c c c c}
        \ja{\texttt{)}、} & \ja{\texttt{)}。} & \texttt{\_)} & \ja{ンの} & \ja{スの} \\ 
        \ja{は} & \ja{のは} & \texttt{\_}\ja{の} & \ja{、}\texttt{2} & \ja{ンは}
        \end{tabular} &
        \begin{tabular}{c c c c c@{}}
        \texttt{li} & \texttt{lo} & \ja{ていく} & \texttt{vi} & \ja{てしまう} \\
        \texttt{hi} & \texttt{0\%} & \texttt{to} & \texttt{no} & \texttt{ta}
        \end{tabular} \\
        \bottomrule
    \end{tabular}
    \caption{Tokens with the highest difference in frequency between tokenizations. The BPE method merges common tokens, such as particles and punctuation, even when they do not form meaningful units. The unigram LM method recovers the units \ja{ていく} and \ja{てしまう}, which are productive components of the Japanese verb conjugation system.}
    \label{tab:max_diff_ja}
\end{table*}

\begin{figure*}[!h]
\small
    \centering
    \begin{subfigure}[t]{0.49\textwidth}
        \includegraphics[scale=0.97]{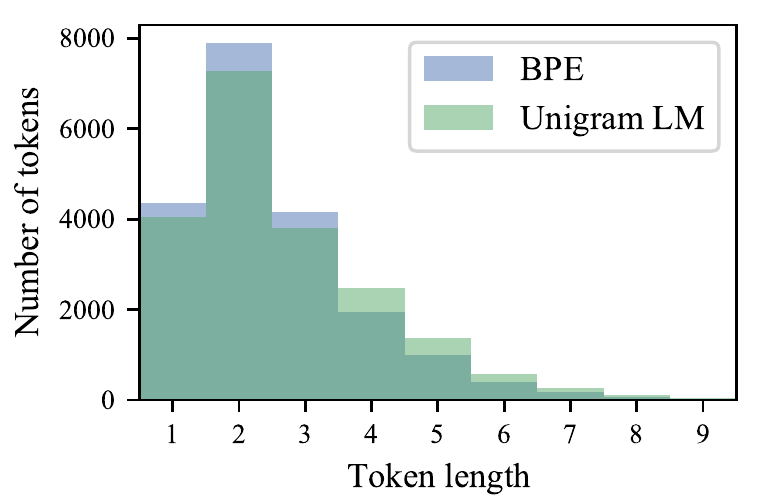}
        \caption{Token length distributions within each vocabulary}
        \label{fig:token_lengths_ja}
    \end{subfigure}
    \begin{subfigure}[t]{0.49\textwidth}
        \includegraphics[scale=0.97]{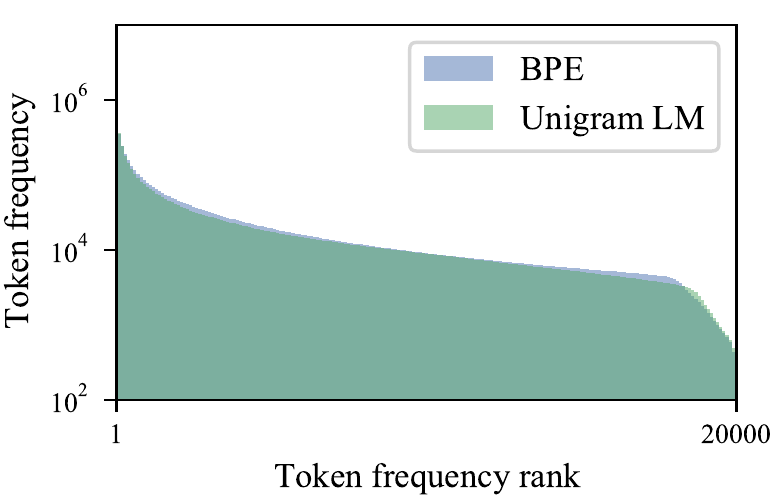}
        \caption{Token frequency profiles over the corpus}
        \label{fig:token_freqs_ja}
    \end{subfigure}
    \caption{Japanese subword vocabulary and corpus profiles. (a) The unigram LM method produces longer tokens, as it does in English. (b) Token frequency profiles resemble those of English, though the effect of the ``dead zone'' is less pronounced.}
\end{figure*}

\newpage

\end{document}